\documentclass[]{fairmeta}
\usepackage{xspace}
\definecolor{cvprblue}{rgb}{0.21,0.49,0.74}

\makeatletter
\DeclareRobustCommand\onedot{\futurelet\@let@token\@onedot}
\def\@onedot{\ifx\@let@token.\else.\null\fi\xspace}

\makeatother

\usepackage{booktabs}
\usepackage{multirow}
\usepackage{colortbl}
\usepackage{adjustbox}
\usepackage{amsfonts}
\usepackage[table,xcdraw]{xcolor}
\usepackage[normalem]{ulem}
\usepackage{enumitem}

\usepackage[utf8]{inputenc} 
\usepackage[T1]{fontenc}    
\usepackage{url}            
\usepackage{booktabs}       
\usepackage{amsfonts}       
\usepackage{nicefrac}       
\usepackage{microtype}      
\usepackage{graphicx}
\usepackage{colortbl}
\usepackage{adjustbox}
\usepackage{multirow}
\usepackage{enumitem}
\usepackage{amsmath}
\definecolor{citecolor}{HTML}{2980b9}
\definecolor{linkcolor}{HTML}{c0392b}

\usepackage{adjustbox}
\newtcolorbox{graylist}{
  enhanced,
  colback=gray!5,    
  colframe=gray!15, 
  boxrule=0pt,      
  arc=0mm,          
  left=1mm, right=1mm, top=1mm, bottom=1mm,
  width=\dimexpr\linewidth-4mm\relax,  
  center,            
  breakable
}

\useunder{\uline}{\ul}{}

\newcommand{\tocite}[1]{\textcolor{red}{[TO CITE]}}

\newcommand{\name}{\textsc{ATLAS}\xspace}

\title{\raisebox{-0.22\height}{\includegraphics[height=1.5em]{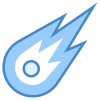}}\textcolor{metablue}{ATLAS}:\hspace{0.22cm} \textcolor{metablue}{A}gen\textcolor{metablue}{t}ic or \textcolor{metablue}{La}tent Vi\textcolor{metablue}{s}ual Reasoning?\\\hspace{3cm} One Word is Enough for Both\vspace{0.2cm}}

\abstract{
Visual reasoning, often interleaved with intermediate visual states, has emerged as a promising direction in the field. A straightforward approach is to directly generate images via unified models during reasoning, but this is computationally expensive and architecturally non-trivial. Recent alternatives include agentic reasoning through code or tool calls, and latent reasoning with learnable hidden embeddings. However, agentic methods incur context-switching latency from external execution, while latent methods lack task generalization and are difficult to train with autoregressive parallelization. To combine their strengths while mitigating their limitations, we propose \textcolor{metablue}{\textbf{\name}}, a framework in which a single discrete ``word'', termed as a functional token, serves both as an agentic operation and a latent visual reasoning unit. Each functional token is associated with an internalized visual operation, yet requires no visual supervision and remains a standard token in the tokenizer vocabulary, which can be generated via next-token prediction. This design avoids verbose intermediate visual content generation, while preserving compatibility with the vanilla scalable SFT and RL training, without architectural or methodological modifications. To further address the sparsity of functional tokens during RL, we introduce Latent-Anchored GRPO (LA-GRPO), which stabilizes the training by anchoring functional tokens with a statically weighted auxiliary objective, providing stronger gradient updates. Extensive experiments and analyses demonstrate that \name achieves superior performance on challenging benchmarks while maintaining clear interpretability. We hope \name offers a new paradigm inspiring future visual reasoning research.}

\author[1,2]{Ziyu Guo}
\author[1]{Rain Liu}
\author[2]{Xinyan Chen}
\author[2]{Pheng-Ann Heng}

\affiliation[1]{Meta AI} 
\affiliation[2]{The Chinese University of Hong Kong}

\vspace{7mm}
\metadata[Project Page]{\url{https://atlas-oneword.github.io}}
\correspondence{\email{guoziyu86@gmail.com}}
\vspace{1.1cm}

\begin{document}

\maketitle

\abstract{
Visual reasoning, often interleaved with intermediate visual states, has emerged as a promising direction in the field. A straightforward approach is to directly generate images via unified models during reasoning, but this is computationally expensive and architecturally non-trivial. Recent alternatives include agentic reasoning through code or tool calls, and latent reasoning with learnable hidden embeddings. However, agentic methods incur context-switching latency from external execution, while latent methods lack task generalization and are difficult to train with autoregressive parallelization. To combine their strengths while mitigating their limitations, we propose \textcolor{metablue}{\textbf{\name}}, a framework in which a single discrete ``word'', termed as a functional token, serves both as an agentic operation and a latent visual reasoning unit. Each functional token is associated with an internalized visual operation, yet requires no visual supervision and remains a standard token in the tokenizer vocabulary, which can be generated via next-token prediction. This design avoids verbose intermediate visual content generation, while preserving compatibility with the vanilla scalable SFT and RL training, without architectural or methodological modifications. To further address the sparsity of functional tokens during RL, we introduce Latent-Anchored GRPO (LA-GRPO), which stabilizes the training by anchoring functional tokens with a statically weighted auxiliary objective, providing stronger gradient updates. Extensive experiments and analyses demonstrate that \name achieves superior performance on challenging benchmarks while maintaining clear interpretability. We hope \name offers a new paradigm inspiring future visual reasoning research.}

\vspace{3mm}
\begin{figure*}[t!]
\vspace{-0.3cm}
    \centering
    \includegraphics[width=\textwidth]{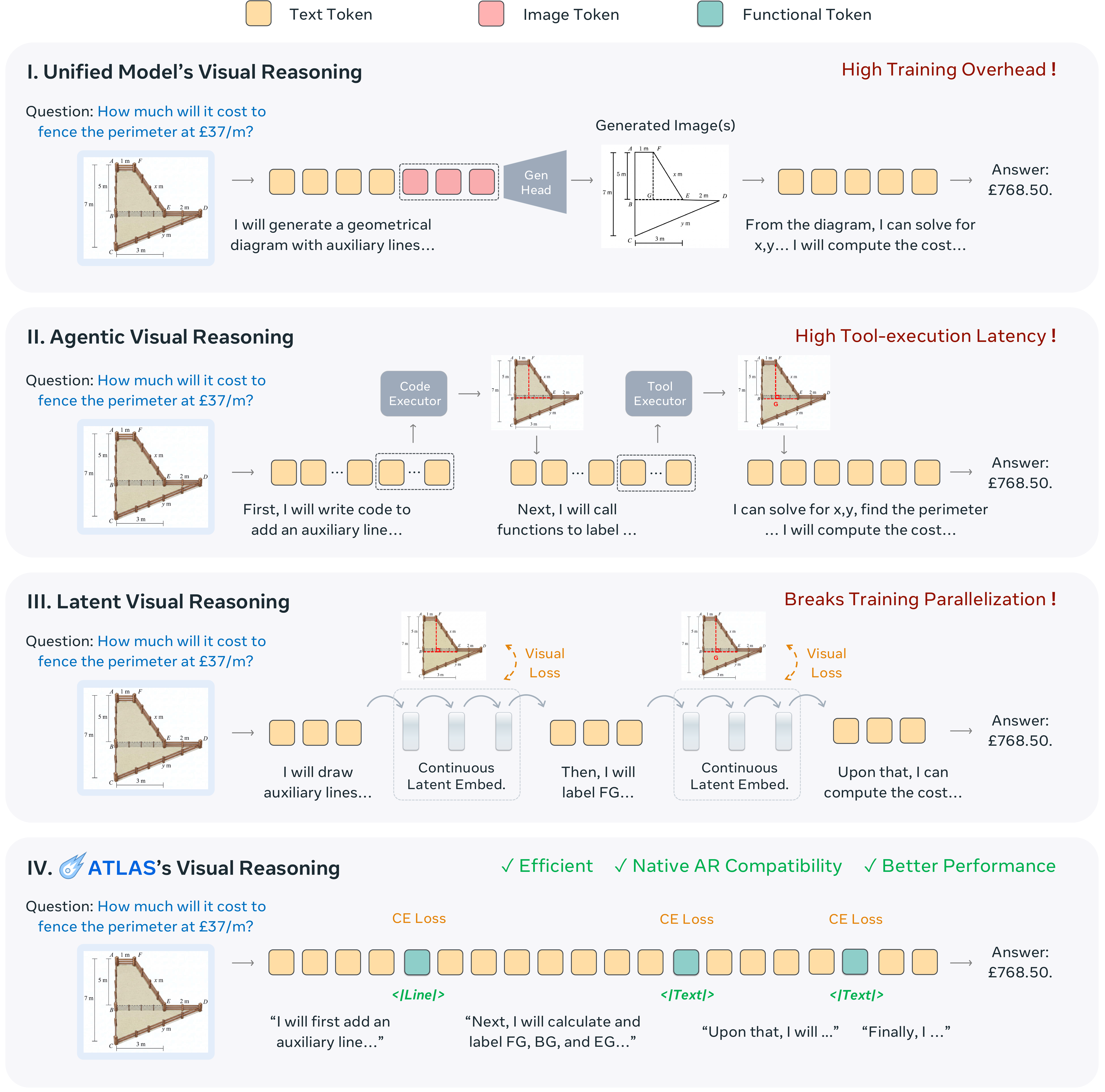}
    \caption{\textbf{Comparison of Visual Reasoning Paradigms.}
    \textbf{I}: Unified models generate intermediate pixel-level images.
    \textbf{II}: Agentic methods rely on external code or tool execution.
    \textbf{III}: Latent methods conduct intermediate reasoning through latent embeddings.
    \textbf{IV}: \textcolor{metablue}{\name} briges agentic and latent visual reasoning through discrete vocabulary functional tokens within the standard autoregressive generation loop, more efficient and effective.}
    \label{fig:intro}
\vspace{-0.1cm}
\end{figure*}

\section{Introduction}
\vspace{3mm}

\label{sec:intro}
The rapid evolution of Vision-Language Models (VLMs)~\cite{bai2025qwen3,an2025llava,bai2025qwen2,seed2026seed1,team2024gemini,li2024llava-ov} has advanced multimodal intelligence from perception toward reasoning~\cite{jiang2025mme}. In these tasks, purely textual reasoning is often insufficient, as problem solving frequently requires intermediate visual analysis~\cite{shao2024visual,zhao2025unified,chern2024anole}. This capability, commonly studied as interleaved visual reasoning, involves generating, perceiving, and using intermediate visual states to guide subsequent inference~\cite{chen2025mint,qiao2025v,su2025pixel}. For instance, game solving may require updating the board state after each operation, while geometry solving may require constructing auxiliary lines to reveal hidden relations~\cite{hu2024visual,zhang2024mathverse}. Despite strong progress in direct visual understanding, current VLMs still remain limited in this dynamic visual reasoning process.

Unified models~\cite{deng2025emerging,li2025imaginereasoningspacemultimodal,zhao2025unified,liu2025tuna,wu2024janus,xie2024show} provide a straightforward solution by explicitly generating pixel-level images, as illustrated in Fig.~\ref{fig:intro}\textcolor{metablue}{I}. This paradigm is intuitive: the model externalizes intermediate visual representations in the same modality as the input. However, generating new images introduces substantial inference cost and training difficulty. The model must allocate significant capacity to image decoding and re-encoding, and requires non-trivial framework-level architectural designs, which often necessitates pre-training from scratch.

To better preserve the standard VLM architecture, existing methods explore two alternative routes. First, \textit{\textbf{agentic visual reasoning}}~\cite{gupta2023visual,hu2024visual,suris2023vipergpt} in Fig.~\ref{fig:intro}\textcolor{metablue}{.II}, treats the VLM as a high-level controller that generates code or tool calls to manipulate the visual input through external modules.
Although its computational overhead is lower than that of generating full intermediate images, it still often requires verbose code or tool-call formulations even for simple visual operations, increasing output length and inference latency.
Second, \textit{\textbf{latent reasoning}}~\cite{wang2025monet,li2025latent,qin2025chain} in Fig.~\ref{fig:intro}\textcolor{metablue}{.III}, performs intermediate reasoning in hidden representations rather than generating images or long textual operations. 
However, the supervision signals for latent embeddings are derived from a specific range of tasks, limiting their generalization to broader domains. More critically, they introduce recurrent latent dependencies~\cite{hao2024training}, which break the compatibility with standard parallel training and substantially increase training cost.

In this paper, we propose \textbf{\textcolor{metablue}{\name}}, a framework in which only a single functional ``word'' serves as both an agentic operation and a latent reasoning unit, as illustrated in Fig.~\ref{fig:intro}\textcolor{metablue}{.IV}. The key idea of \name\ is to represent each visual operation as a standard discrete token in the tokenizer vocabulary, such as zooming into a region, constructing auxiliary lines, drawing shapes, adding arrows, or inserting textual labels. These tokens are generated through ordinary next-token prediction within the same sequence as natural language tokens, rather than being modeled as continuous latent states outside the autoregressive sequence. 

Compared with agentic methods, \name\ provides a compact and efficient interface that internalizes complex code generation, tool calling, and external execution into a single token. Compared with latent methods, \name\ maintains a standard autoregressive generation loop without any visual supervision, preserving compatibility with existing supervised fine-tuning (SFT) and reinforcement learning (RL) frameworks, enabling efficient parallel training with scalability to larger-size models and data.
It is also worth noting that these functional tokens do not require image-level supervision. Instead, they are optimized with the standard cross-entropy (CE) objective over token sequences, allowing the model to learn from the reasoning context by iteself when and how to invoke them as effective visual operations.

We adopt a two-stage training recipe for \name. First, to provide a reliable cold start for using functional tokens, we curate a new dataset, \name-178K, covering over 40 visual reasoning tasks collected and reformulated from existing efforts~\cite{qiao2025v}. Each example is annotated with functional-token trajectories that specify the desired visual operations, enabling the model to learn when and how to invoke functional tokens within standard autoregressive generation. 
On top of this, we apply RL to enhance visual reasoning through outcome-driven optimization. Thanks to our designs that functional tokens are represented as ordinary vocabulary tokens, \name can be optimized directly with standard GRPO~\cite{shao2024deepseekmath}, without introducing customized training modifications~\cite{liu2025flow,xue2025dancegrpo}. We leverage a diverse reward ensemble that jointly encourages answer correctness, valid functional-token usage, and coherent reasoning behavior, which already yields improvements over the SFT model.

However, during RL training, we observe a critical ``gradient dilution'' issue: the sparse functional tokens responsible for visual reasoning are overwhelmed by the much larger number of ordinary text tokens, leading to insufficient optimization. To mitigate this, we introduce Latent-Anchored GRPO (LA-GRPO), which augments the standard GRPO objective with a statically weighted token-level auxiliary loss anchored on the functional-token vocabulary. This auxiliary objective provides a persistent learning signal for functional tokens, yielding consistent performance gains across reasoning tasks.

Our contributions are summarized as follows:

\begin{itemize}
    \item We propose \name, a visual reasoning framework that represents visual operations as discrete functional tokens in the standard vocabulary, avoiding verbose intermediate visual states, while preserving compatibility with scalable autoregressive training.

    \item We identify gradient dilution for sparse functional tokens during training and propose LA-GRPO, a token-anchored objective that strengthens functional-token optimization.

    \item We show that \name\ enables compact single-token visual reasoning, achieving strong performance on challenging benchmarks with substantially reduced overhead.
\end{itemize}
\begin{figure*}[t!]
    \centering
    \includegraphics[width=\textwidth]{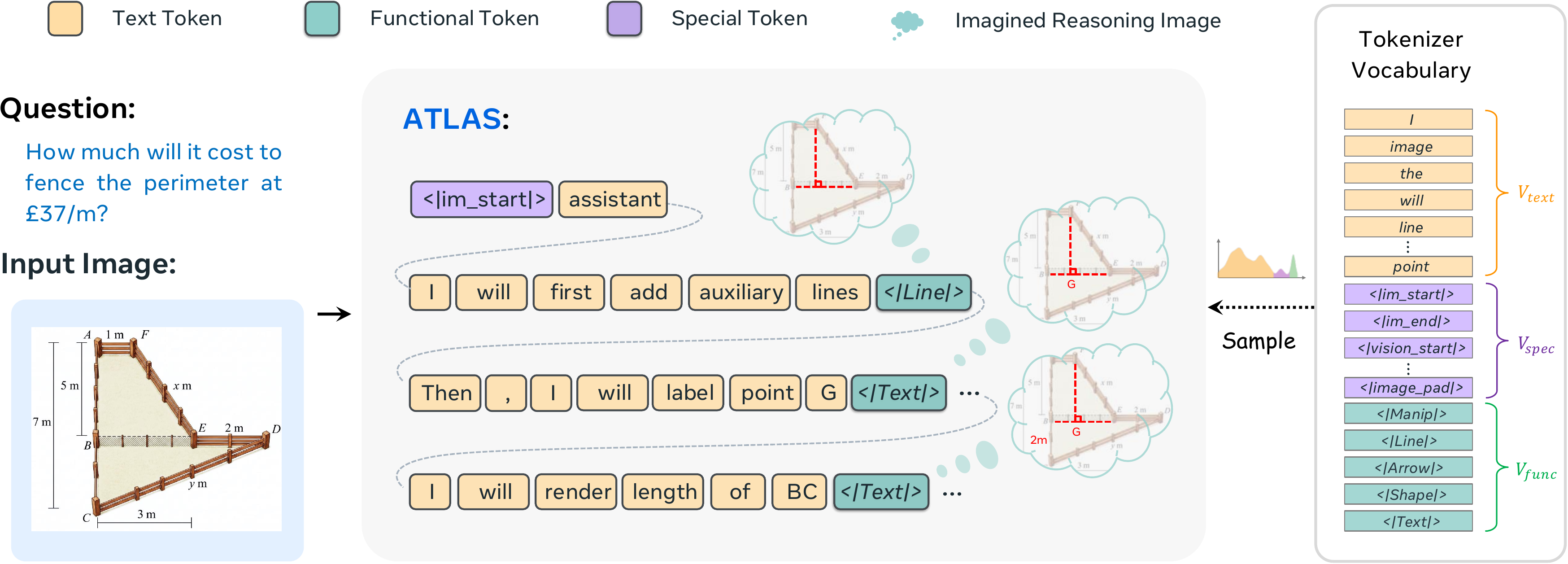}
    \caption{\textbf{Overall Pipeline of \name.} \name represents visual operations as functional tokens within the standard autoregressive sequence, enabling the model to perform visual reasoning without generating intermediate images or invoking external tools.}
    \label{fig:pipeline}
\end{figure*}

\section{\textcolor{metablue}{\name}}
\label{sec:method}

In this section, we present \name, a framework that bridges agentic and latent visual reasoning through discrete functional tokens. We first introduce the overall model architecture in Sec.~\ref{model_arch}, including the design of functional tokens within the autoregressive sequence. We then describe the training paradigm in Sec.~\ref{training_paradigm}, which consists of an SFT on the curated \name-178K dataset followed by a standard RL with GRPO~\cite{shao2024deepseekmath}. Finally, in Sec.~\ref{sec:la_grpo}, we present the proposed LA-GRPO objective for enhanced functional-token optimization.

\subsection{Model Architecture}
\label{model_arch}

Building upon standard autoregressive architectures~\cite{bai2025qwen2,llavanext2024,bai2025qwen3}, \name formulates visual reasoning as next-token prediction by representing visual operations as discrete learnable functional tokens in the tokenizer vocabulary. We instantiate \name\ with Qwen2.5-VL~\cite{bai2025qwen2} and add five functional tokens, each corresponding to an internalized operation. Generated like ordinary words within the same autoregressive sequence, these tokens provide a compact and interpretable interface for active perception and visual construction, while avoiding external tool execution, pixel-level intermediate supervision, and recurrent latent dependencies. This preserves compatibility with existing VLM pipelines and supports efficient parallel training.

\paragraph{Taxonomy of Functional Tokens.}
To internalize visual operations into the reasoning process, we expand the standard vocabulary $\mathcal{V}$ with a compact set of functional tokens. Formally, the full vocabulary is defined as
\[
\mathcal{V} = \mathcal{V}_{text} \cup \mathcal{V}_{spec} \cup \mathcal{V}_{func},
\]
where $\mathcal{V}_{text}$ denotes natural language tokens, $\mathcal{V}_{spec}$ denotes the original special tokens of the VLM (\textit{e.g.}, \texttt{<im\_start>}, \texttt{<image\_pad>}), and
\[
\mathcal{V}_{func} = \{ \texttt{<|Manip|>}, \texttt{<|Shape|>}, \texttt{<|Line|>}, \texttt{<|Arrow|>}, \texttt{<|Text|>} \}
\]
denotes the five proposed functional tokens.
We intentionally keep $\mathcal{V}_{func}$ compact to avoid excessive perturbation to the original token distribution of the base model. Instead of introducing many task-specific tokens, we abstract common visual operations into a small set of general categories. For instance, bounding boxes, masks, cropping, and zooming can all be represented by the generalized region-based token \texttt{<|Shape|>}. As summarized in Tab.~\ref{tab:functional_tokens}, each functional token corresponds to a high-level visual operation that can support multi-step reasoning.
This taxonomy is not intended to be exhaustive. Rather, it provides a simple and effective template for internalizing visual operations as discrete tokens. Future work can naturally extend the functional-token vocabulary to cover more diverse operations and scenarios.

\begin{table}[t!]
    \centering
    \small
    \caption{\textbf{Taxonomy of Functional Tokens.} The imagined reasoning images illustrate the internal visual reasoning states, which will not be generated or supervised in the sequence.}
    \label{tab:functional_tokens}
    \begin{tabular}{@{}>{\centering\arraybackslash}m{0.18\textwidth}
                    >{\raggedright\arraybackslash}m{0.5\textwidth}
                    >{\centering\arraybackslash}m{0.26\textwidth}@{}}
    \toprule
    \textbf{Functional Token}
    & \textbf{Description}
    & \textbf{Imagined Reasoning Image} \\
    \cmidrule(lr){1-1}\cmidrule(lr){2-2}\cmidrule(l){3-3}

    \texttt{<|Manip|>}
    & Enhances image through denoising, sharpening, filtering, cropping, or zoom-in based inspection.
    & \includegraphics[width=0.25\textwidth]{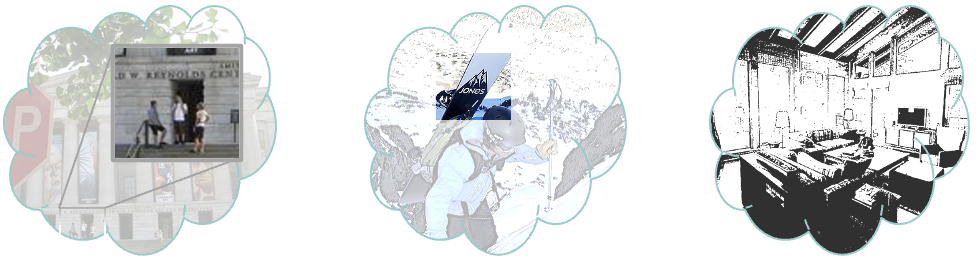} \\[4pt]

    \texttt{<|Shape|>}
    & Marks regions for visual grounding, highlighting, or area reasoning.
    & \includegraphics[width=0.25\textwidth]{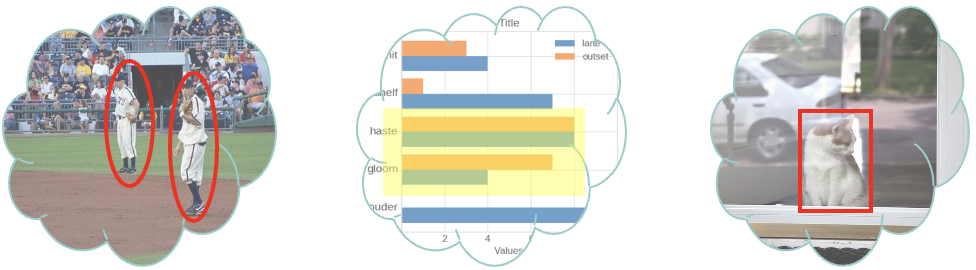} \\[4pt]

    \texttt{<|Line|>}
    & Adds lines for geometric construction, visual separation, structural cues, or emphasis.
    & \includegraphics[width=0.25\textwidth]{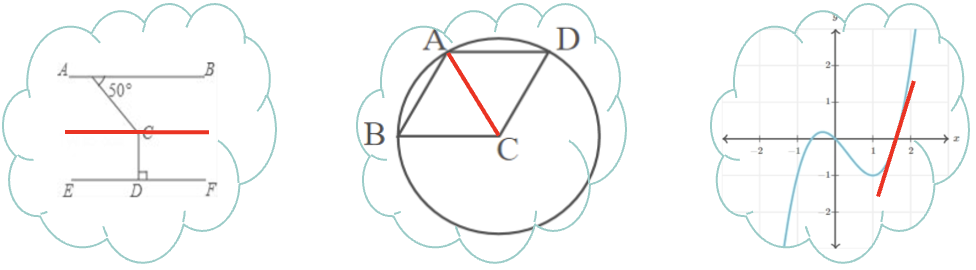} \\[4pt]

    \texttt{<|Arrow|>}
    & Indicates direction, motion, force, flow, causal relation, or highlights key visual elements.
    & \includegraphics[width=0.25\textwidth]{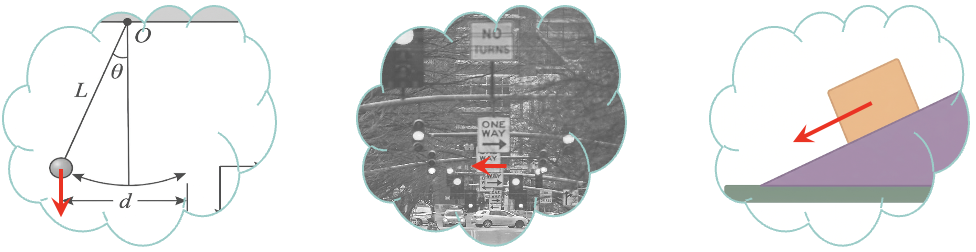} \\[4pt]

    \texttt{<|Text|>}
    & Adds symbolic labels, numerical values, or textual notes to support multi-step reasoning.
    & \includegraphics[width=0.25\textwidth]{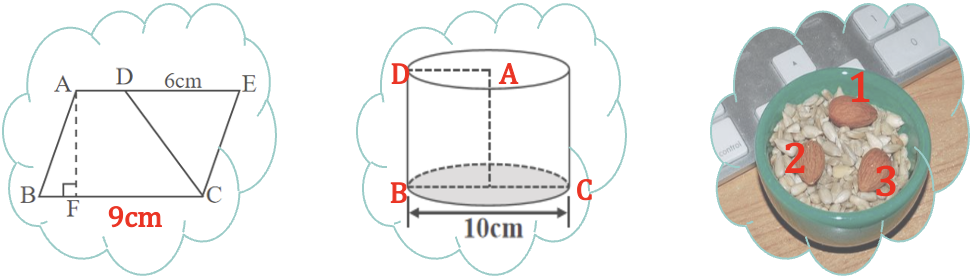} \\

    \bottomrule
    \end{tabular}
\end{table}

\paragraph{Unified Sequence Modeling.}
Unlike agentic approaches that pause generation to call external modules, or latent methods that produce continuous hidden embeddings, \name\ keeps the entire reasoning process within a single discrete autoregressive sequence. Given a multimodal input context $\mathbf{x}$, the model predicts an output sequence:
\[
\mathbf{y} = \{y_1, \dots, y_T\},\ \ \ \ \text{where}\ \ y_t \in \mathcal{V} = \mathcal{V}_{text} \cup \mathcal{V}_{spec} \cup \mathcal{V}_{func}
\] When a functional token $y_t \in \mathcal{V}_{func}$ is predicted, it is treated as an ordinary sequence token while serving as an internal reasoning unit that specifies the type of visual operation needed at the current step. For example, \texttt{<|Line|>} indicates that the model should reason with an auxiliary line, while \texttt{<|Text|>} indicates that symbolic labels or numerical annotations may be useful for the subsequent derivation. This formulation preserves the explicitness and interpretability of agentic reasoning, while avoiding the latency of tool execution and the cost of pixel-level image generation.
Importantly, functional tokens do not require any image-level supervision. Instead, they are optimized with the same cross-entropy (CE) objective as ordinary text tokens:
\[
\mathcal{L}_{func}
= - \sum_{y_t \in \mathcal{V}_{func}}
\log p_\theta(y_t \mid \mathbf{x}, y_{<t}).
\]

Through token-level supervision, the model learns from the surrounding reasoning context when and how to invoke functional tokens as effective visual operations. For example, when the reasoning context states, \textit{``Now I will add an auxiliary height to ...''}, the next functional token can be \texttt{<|Line|>}, encouraging the model to associate such geometric construction intent with the corresponding functional token. Since all reasoning units remain within the autoregressive sequence, \name\ is fully compatible with scalable next-token training and inference pipelines.

\subsection{Two-stage Training Recipe}
\label{training_paradigm}

We train \name\ in two stages. First, we curate \name-178K, an SFT dataset tailored to our visual reasoning paradigm with functional tokens. This provide a cold start for functional-token invocation and improved interleaved visual reasoning. Second, we apply standard GRPO for RL, further enhancing reasoning performance through reward-guided optimization.

\paragraph{Stage 1: SFT with \name-178K.}
We construct \name-178K to provide supervised reasoning trajectories for the SFT stage. Specifically, it is constructed through the following three steps:

\begin{enumerate}
  \item \textit{\textbf{Source Data and Token Extraction:}} We start from the publicly released preview subset of \texttt{V-Interaction-400K}~\cite{qiao2025v}, which provides image-construction code paired with visual reasoning problems, making it suitable for deriving functional-token supervision. We parse the original code and extract visual operations that can be naturally mapped to our functional-token space, including line drawing, text annotation, shape drawing, visual refinement, cropping, and other visually grounded transformations. We then filter the extracted samples and retain 138K high-quality examples covering over 40 tasks for functional-token trajectory construction.
  
  \item \textit{\textbf{Trajectory Construction and Polishing:}}
  After extracting the mapped operations, we convert them into reasoning trajectories with functional tokens. For each functional step, we insert a predefined transition template so that the functional token appears as an explicit part of the reasoning process. Since directly templated trajectories can be overly rigid, we further use Gemini-2.5-Pro~\cite{team2024gemini} to polish them into more natural reasoning text while preserving the original semantics and functional-token order.
  
  \item \textit{\textbf{Perception Preservation:}}
  To preserve the model's low-level perceptual ability, we also include \texttt{V-Perception-40K}~\cite{qiao2025v} during SFT. This part of data does not contain functional tokens, but provides complementary supervision for fine-grained visual understanding and helps reduce catastrophic forgetting during fine-tuning.
\end{enumerate}

With this dataset, we train the model using the vanilla CE loss~\cite{mao2023cross}, updating all tokens in the sequence and enabling the model to learn valid functional-token invocation from context.

\paragraph{Stage 2: Standard RL with GRPO.}
While SFT provides a cold start for functional-token usage, complex multi-step reasoning further requires the model to decide when such operations are useful for reaching the correct answer. Thanks to the compatibility of \name with standard autoregressive generation, we can directly adopt GRPO without introducing customized training procedures.
Given a query $q$, the policy $\pi_{\theta}$ samples a group of $G$ outputs $\{o_1,\dots,o_G\}$. We define a composite reward $r(o)$ that encourages answer correctness, effective functional-token usage, and valid formatting, while penalizing overly long responses and excessive token invocation:
\[
r(o) = \lambda_{\text{acc}} r_{\text{acc}}
+ \lambda_{\text{func}} r_{\text{func}}
+ \lambda_{\text{fmt}} r_{\text{fmt}}
- \lambda_{\text{len}} p_{\text{len}}
- \lambda_{\text{spam}} p_{\text{spam}},
\]
where each $\lambda$ controls the corresponding reward or penalty term defined as follows:

\begin{itemize}
    \item \textbf{Answer Accuracy ($r_{\text{acc}}$):} Evaluates whether the final answer is correct. We use exact string matching and mathematical equivalence checking when applicable. The reward is $1$ for a correct answer and $0$ otherwise.

    \item \textbf{Functional Token Usage ($r_{\text{func}}$):} To encourage the meaningful usage of functional tokens and prevent reward hacking, we implement a strict conditional reward mechanism. The functional-token reward is granted only if the model invokes at least one functional token and successfully get the correct final answer.

    \item \textbf{Format Adherence ($r_{\text{fmt}}$):} Ensures that the final answer follows the required format for reliable parsing. It yields $1$ if the required format is satisfied and $0$ otherwise.

    \item \textbf{Length Penalty ($p_{\text{len}}$):} Discourages overly verbose responses. If the output length $L(o)$ exceeds a predefined threshold $L_{max}$, we apply a linear penalty within a fixed buffer range, capped by a maximum penalty value.

    \item \textbf{Token Overuse Penalty ($p_{\text{spam}}$):} Prevents the model from repeatedly generating functional tokens only to exploit the usage reward. Let $N_{func}$ denote the number of functional tokens in the output. If $N_{func}$ exceeds a threshold $\tau_{spam}$, we apply a bounded linear penalty for excessive usage. 
\end{itemize}

This reward design reflects a simple principle: functional tokens should be encouraged only when they support effective reasoning. The standard GRPO objective then optimizes the policy according to the relative advantage within the sampled group:
\[
\mathcal{L}_{GRPO}(\theta) =
- \frac{1}{G} \sum_{i=1}^{G}
\left(
\frac{\pi_{\theta}(o_i \mid q)}{\pi_{ref}(o_i \mid q)}
\right)^{\beta}
\hat{A}_i
- \beta \mathbb{D}_{KL}(\pi_{\theta} || \pi_{ref}),
\]
where $\hat{A}_i$ is the advantage computed from the group rewards $r(o_i)$, and $\beta$ controls the KL penalty.

\begin{figure*}[t!]
    \centering
    \includegraphics[width=\textwidth]{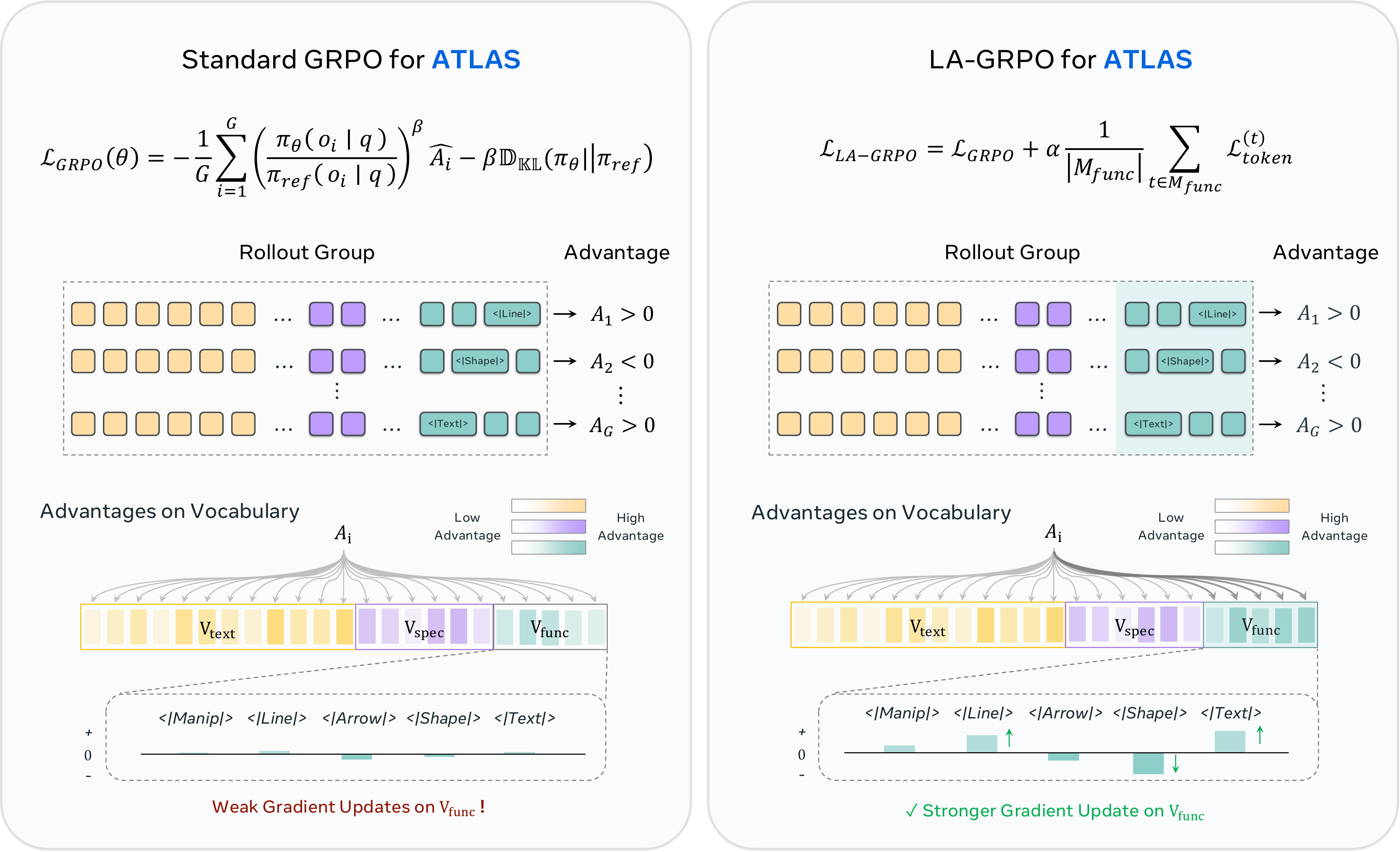}
    \caption{\textbf{Latent-Anchored GRPO.} Standard GRPO provides sequence-level advantages to all generated tokens, which can dilute the learning signal for sparse functional tokens. LA-GRPO adds a token-level auxiliary objective on $\mathcal{V}_{func}$ to stabilize functional-token optimization.}
    \label{fig:lagrpo}
\end{figure*}

\subsection{LA-GRPO}
\label{sec:la_grpo}

Directly applying standard GRPO to \name suffers from a ``gradient dilution'' issue. As illustrated in Fig.~\ref{fig:lagrpo}, standard GRPO assigns a sequence-level advantage to each rollout response and propagates this signal to all generated tokens. However, functional tokens occupy only a very small portion of the sequence. In \name trajectories, an average response contains 203.7 generated tokens, but only 4.8 of them are functional tokens, corresponding to a ratio of 2.3\%. As a result, the learning signal for these sparse but important visual-operation tokens is easily diluted by the much larger number of ordinary text tokens. This weakens updates on $\mathcal{V}_{func}$ and may cause the model to underuse functional tokens or learn unstable behaviors such as token spamming.

To address this issue, we propose Latent-Anchored GRPO (LA-GRPO). The key idea is to keep the original sequence-level GRPO objective unchanged, while adding a functional-token anchor that explicitly strengthens optimization on $\mathcal{V}_{func}$. Concretely, for each sampled rollout, we identify the positions where functional tokens appear and apply an additional token-level auxiliary objective only to these positions. This auxiliary term reuses the rollout advantage from GRPO, but concentrates the update on functional tokens such as \texttt{<|Line|>}, \texttt{<|Shape|>}, and \texttt{<|Text|>}. In this way, LA-GRPO preserves the global reward-driven optimization of standard GRPO while providing a stronger and more persistent learning signal for the tokens responsible for internalized visual operations.

Specifically, for each rollout $o_i=\{y_1,\dots,y_T\}$, we collect the positions of functional tokens:
\[
M_{\mathrm{func}}(o_i)=\{t \mid y_t \in \mathcal{V}_{func}\}.
\]
For each $t\in M_{\mathrm{func}}(o_i)$, we define a token-level clipped surrogate loss:
\[
\mathcal{L}_{\mathrm{token}}^{(t)}
=
-\min\left(
\rho_{i,t}\hat{A}_i,\,
\mathrm{clip}(\rho_{i,t},1-\epsilon,1+\epsilon)\hat{A}_i
\right),
\quad
\rho_{i,t}
=
\frac{\pi_{\theta}(y_t \mid q,y_{<t})}
{\pi_{\theta_{\mathrm{old}}}(y_t \mid q,y_{<t})}.
\]
This objective anchors the group-level advantage directly to functional-token positions, producing stronger updates on sparse visual-operation tokens. Then, the final objective is:
\[
\mathcal{L}_{\mathrm{LA\text{-}GRPO}}
=
\mathcal{L}_{\mathrm{GRPO}}
+
\alpha
\frac{1}{|M_{\mathrm{func}}|}
\sum_{t \in M_{\mathrm{func}}}
\mathcal{L}_{\mathrm{token}}^{(t)},
\]
where $\alpha$ controls the anchor strength and the normalization stabilizes the auxiliary loss scale. Unlike standard GRPO, which spreads the advantage signal across the whole vocabulary, LA-GRPO explicitly reinforces $\mathcal{V}_{func}$ while retaining sequence-level reward optimization. This yields stronger gradients for functional tokens, stabilizes their invocation, and improves visual reasoning without modifying the autoregressive training pipeline.

\section{Experiments}
\subsection{Implementation Details}
\paragraph{Training Details.} 
We adopt Qwen2.5-VL-7B~\cite{bai2025qwen2} as our base model. During the SFT stage, we freeze the vision encoder and solely update the visual projector and the language model. For the RL stage, we continue to freeze the vision encoder while optimizing the aligner and language model for 1 epoch.

\begin{table*}[t!]
    \centering
    \small
    \caption{\textbf{Mapping from Code Operations to Functional Tokens.} We parse visual operations from image-construction code and map them into a compact functional-token space.}
    \label{tab:code2func}
    \renewcommand{\arraystretch}{1.05}
    \setlength{\tabcolsep}{8pt}
    \begin{tabular}{@{}>{\centering\arraybackslash}p{0.2\textwidth}
                    >{\raggedright\arraybackslash}p{0.38\textwidth}@{}}
    \toprule
    \textbf{Functional Token}
    & \textbf{Representative Code Patterns} \\
    \cmidrule(lr){1-1}\cmidrule(l){2-2}

    \texttt{<|Manip|>}
    & \texttt{np.pad(...)} \\
    & \texttt{cv2.blur(...)} \\
    & \texttt{cv2.GaussianBlur(...)} \\
    & \texttt{scipy.signal.convolve(...)} \\
    & \texttt{cv2.filter2D(...)} \\[4pt]
    \cmidrule(lr){1-1}\cmidrule(l){2-2}

    \texttt{<|Line|>}
    & \texttt{plt.plot([x1, x2], [y1, y2], ...)} \\
    & \texttt{ax.plot([x1, x2], [y1, y2], ...)} \\
    & \texttt{cv2.line(img, pt1, pt2, ...)} \\[4pt]
    \cmidrule(lr){1-1}\cmidrule(l){2-2}

    \texttt{<|Arrow|>}
    & \texttt{plt.arrow(x, y, dx, dy, ...)} \\
    & \texttt{ax.arrow(x, y, dx, dy, ...)} \\
    & \texttt{cv2.arrowedLine(img, pt1, pt2, ...)} \\[4pt]
    \cmidrule(lr){1-1}\cmidrule(l){2-2}

    \texttt{<|Shape|>}
    & \texttt{plt.fill(x, y, ...)} \\
    & \texttt{ax.add\_patch(Circle(...))} \\
    & \texttt{ax.add\_patch(Rectangle(...))} \\
    & \texttt{cv2.rectangle(img, pt1, pt2, ...)} \\
    & \texttt{cv2.polylines(img, pts, ...)} \\
    & \texttt{img[y1:y2, x1:x2]} \\
    & \texttt{PIL.Image.crop((x1, y1, x2, y2))} \\
    & \texttt{cv2.resize(crop, ...)} \\
    & \texttt{torchvision.transforms.Resize(...)} \\[4pt]
    \cmidrule(lr){1-1}\cmidrule(l){2-2}

    \texttt{<|Text|>}
    & \texttt{plt.text(x, y, s, ...)} \\
    & \texttt{ax.text(x, y, s, ...)} \\
    & \texttt{cv2.putText(img, text, org, ...)} \\
    \bottomrule
    \end{tabular}
\end{table*}

\begin{table*}[!t]
\caption{\textbf{Performance Comparison on Visual Reasoning.} We compare five groups of VLMs with our \name models on three challenging benchmarks.}
    \centering
    \small
    \renewcommand{\arraystretch}{1.08}
    \setlength{\tabcolsep}{4.5pt}
    \begin{adjustbox}{width=\linewidth}
    \begin{tabular}{p{0.32\linewidth} c c cccccccc}
    \toprule
    \multirow{2}{*}{\textbf{Method}} 
    & \multirow{2}{*}{\textbf{V*}} 
    & \multirow{2}{*}{\textbf{WeMath}}
    & \multicolumn{8}{c}{\textbf{BLINK}} \\
    \cmidrule(lr){4-11}
    & & & \textbf{Avg.} & \textbf{Art.} & \textbf{Count.} & \textbf{Forensic.} & \textbf{IQ} & \textbf{Jigsaw} & \textbf{M-view.} & \textbf{Spatial} \\
    \midrule

    \rowcolor{gray!10}
    \multicolumn{11}{c}{\textit{Closed-source Models}} \vspace{3pt}\\
    GPT-4o~\cite{OpenAI_GPT4o} & 62.8 & 50.6 & 61.0 & 82.9 & 49.2 & 79.5 & 31.3 & 55.3 & 59.4 & 69.2 \\
    Claude-4-Sonnet~\cite{anthropic2025claude4} & 15.2 & 63.0 & 49.9 & 61.5 & 59.2 & 35.6 & 30.0 & 53.3 & 47.4 & 62.2 \\
    Gemini-2.0-Flash~\cite{pichai2024gemini2} & 73.3 & 47.4 & 45.3 & 56.4 & 55.0 & 30.3 & 25.3 & 48.7 & 43.6 & 58.0 \\
    Gemini-2.5-Pro~\cite{comanici2025gemini} & 79.1 & 71.3 & 74.6 & 85.5 & 78.3 & 89.4 & 43.3 & 85.3 & 50.4 & 90.2 \\
    \midrule

    \rowcolor{gray!10}
    \multicolumn{11}{c}{\textit{Standard VLMs}} \vspace{3pt}\\
    Qwen2.5-VL~\cite{bai2025qwen2} & 70.2 & 36.2 & 22.8 & 29.9 & 58.3 & 0.8 & 18.7 & 31.3 & 0.0 & 20.3 \\
    LLaVA-OneVision-7B~\cite{an2025llava} & 75.4 & 23.1 & 36.6 & 47.0 & 43.3 & 25.0 & 20.7 & 38.7 & 33.8 & 47.6 \\
    MiniGPT-v2~\cite{lin2023dual} & 35.6 & 11.0 & 32.8 & 43.6 & 13.3 & 24.2 & 20.3 & 34.7 & 48.9 & 44.8 \\
    Gemma-3-27B~\cite{gemmateam2025gemma3technicalreport} & 62.3 & 31.7 & 32.1 & 42.7 & 37.5 & 21.2 & 16.0 & 33.3 & 31.6 & 42.7 \\
    \midrule
    
    \rowcolor{gray!10}
    \multicolumn{11}{c}{\textit{Unified Models}} \vspace{3pt}\\
    Anole~\cite{chern2024anole} & 25.4	&24.7	&16.4	&31.6	&25.0	&11.7	&14.3	&2.0	&3.0	&27.3\\
    Bagel~\cite{deng2025emerging} & 55.5 & 39.4 & 51.1 & 63.2 & 60.8 & 37.1 & \textbf{30.0} & 57.3 & 39.8 & 69.2 \\
    \midrule

    \rowcolor{gray!10}
    \multicolumn{11}{c}{\textit{Agentic Visual Models}}\vspace{3pt}\\
    Visual CoT~\cite{shao2024visual} & 44.5 & 28.6 & 44.4 & 47.0 & 57.5 & 25.0 & 20.7 & 52.7 & 44.4 & 63.6 \\
    V-Thinker~\cite{qiao2025v} & 41.4	& 32.5	& 35.0	& 26.9	& 43.3	& 19.7	&18.7 &42.0 &51.1	& 43.4 \\
    VTS-V~\cite{bai2025multi} & 74.9 & 42.8 & 51.2 & 62.4 & 61.7 & 32.9 & 28.7 & 56.1 & 49.4 & 67.2 \\
    \midrule

    \rowcolor{gray!10}
    \multicolumn{11}{c}{\textit{Latent Visual Models}} \vspace{3pt}\\
    LVR~\cite{li2025latent} & 77.5 & 41.2 & 49.4 & 59.0 & 60.0 & 35.6 & 25.3 & 52.7 & 48.1 & 65.0 \\
    MCOT~\cite{zhang2023multimodal} & 76.4 & 39.6 & 47.4 & 55.6 & 57.5 & 33.3 & 26.7 & 50.7 & 45.9 & 62.2 \\
    CoVT~\cite{qin2025chain} & 72.8 & 38.1 & 47.9 & 59.0 & 60.0 & 36.4 & 24.0 & 41.3 & 49.6 & 65.0 \\
    Monet~\cite{wang2025monet} & 77.8 & 36.9 & 41.8 & 41.0 & 56.7 & 22.7 & 28.0 & 45.3 & 40.5 & 58.7 \\
    \midrule

    \rowcolor{metablue!8}
    \multicolumn{11}{c}{\textit{Ours}} \vspace{3pt}\\
    \textbf{\textcolor{metablue}{ATLAS}}$_{\textit{SFT}}$ & 77.5	& 28.9	& 46.0	& 50.4	& 59.2	& 26.5	& 26.0	& 54.7	& 48.1	& 57.3 \\
    \textbf{\textcolor{metablue}{ATLAS}}$_{\textit{GRPO}}$ & \textbf{77.9} & 40.3 & 50.5 & 57.3 & 61.7 & 34.1 & 26.0 & \textbf{57.7} & 43.6 & \textbf{70.6} \\
    \textbf{\textcolor{metablue}{ATLAS}}$_{\textit{LA-GRPO}}$ &75.4	&\textbf{45.0}	&\textbf{51.3}	&\textbf{65.0}	&\textbf{62.5}	&\textbf{37.9}	&26.3	&51.3	&\textbf{53.4}	&62.9\\
    \bottomrule
    \end{tabular}
    \end{adjustbox}
    \label{tab:benchmark_results}
\end{table*}

\paragraph{Training Data Details.} 
We parse the original code and extract visual actions that can be naturally mapped to our functional-token space. The detailed correspondences between code actions and functional tokens are summarized in Tab.~\ref{tab:code2func}. Based on that, for each functional step, we insert a predefined transition template so that the functional token appears as an explicit part of the reasoning process. After enhancing by Gemini-2.5-Pro~\cite{comanici2025gemini}, the resulting SFT data contains functional-token reasoning trajectories with logically consistent steps, natural transitions, and correct final answers. For GRPO and LA-GRPO, we use \textit{We-Math 2.0}~\citep{qiao2025we}, \textit{MMK12}~\citep{meng2025mm}, and \textit{ThinkLite}~\citep{wang2025sota}. These datasets cover different visual reasoning scenarios and provide diverse supervision for reinforcement learning.

\paragraph{Evaluation Datasets and Metrics.}
We evaluate \name across a suite of benchmarks, including V*~\cite{wu2024v}, BLINK~\cite{fu2024blink}, and WeMath~\cite{qiao2025we}. We adopt a rigorous automated judging pipeline, where we first employ rule-based scripts to parse the answer from the model outputs and then utilize Qwen3-VL-235B-A22B-Instruct~\cite{bai2025qwen3} as a deterministic LLM-as-a-judge to verify the correctness of the answer and the format.

\begin{figure*}[t!]
    \centering
    \includegraphics[width=0.97\textwidth]{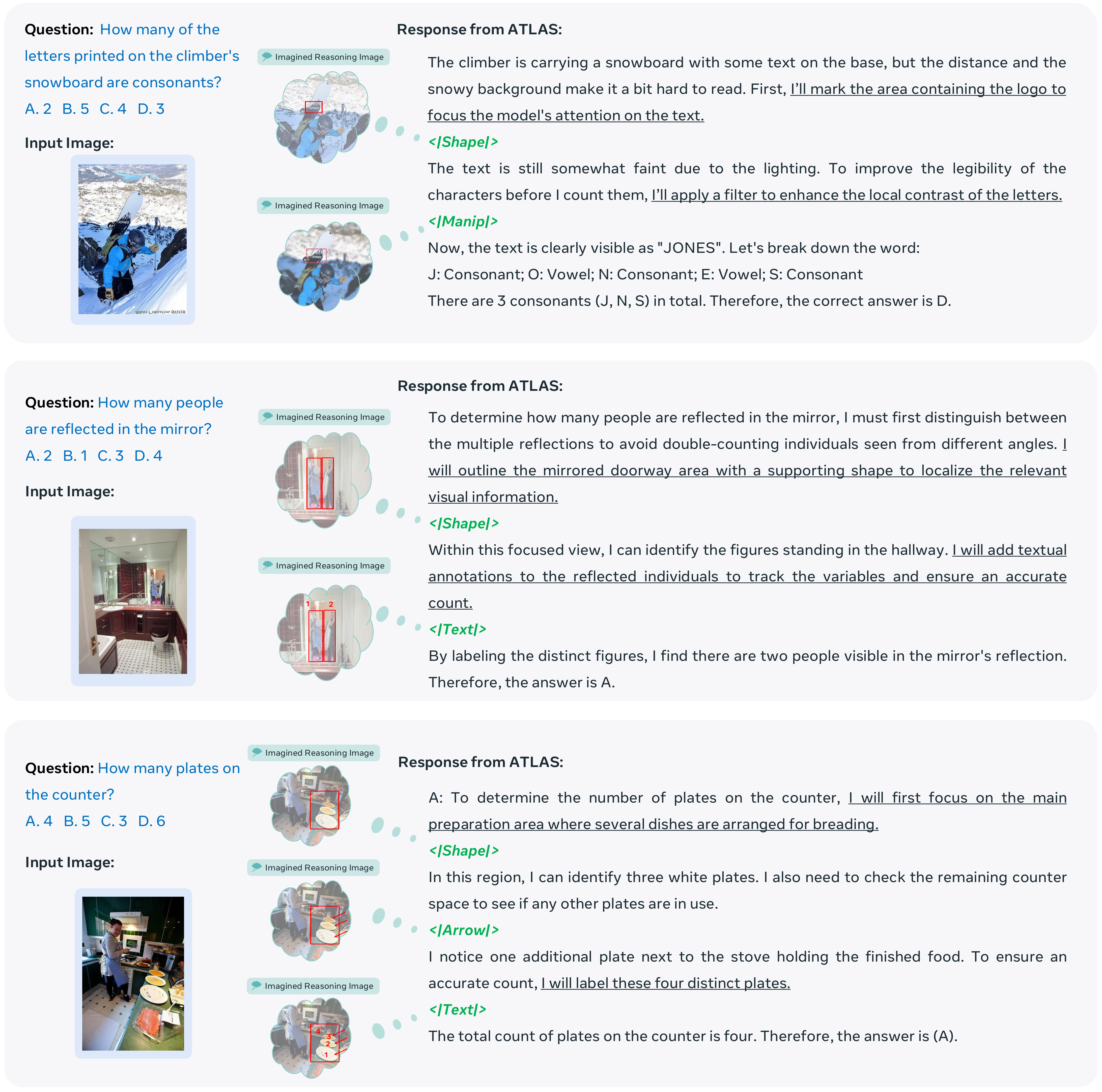}
    \caption{\textbf{Qualitative Examples of \name.} The cloud-shaped regions in the middle illustrate the model's imagined reasoning states for better understanding.}
    \label{fig:qualitative_analysis}
\end{figure*}

\begin{figure*}[t!]
    \centering
    \includegraphics[width=\textwidth]{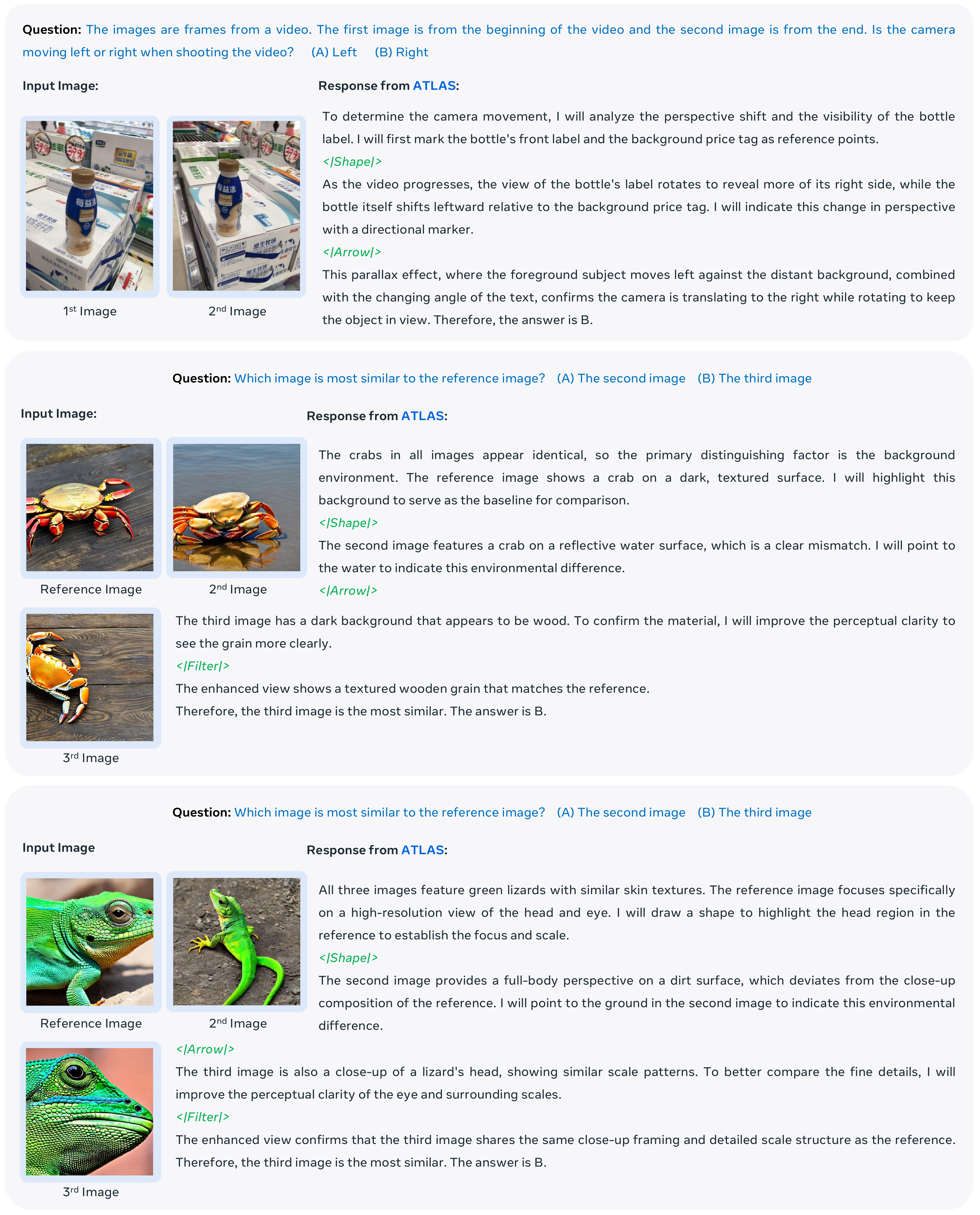}
    \caption{\textbf{Qualitative Examples of \name.} We show more visual reasoning examples where functional tokens help localize relevant regions, indicate directions, refine visual evidence, and support multi-step reasoning.}
    \label{fig:additional_qualitative}
\end{figure*}

\begin{figure*}[t!]
    \centering
    \includegraphics[width=0.97\textwidth]{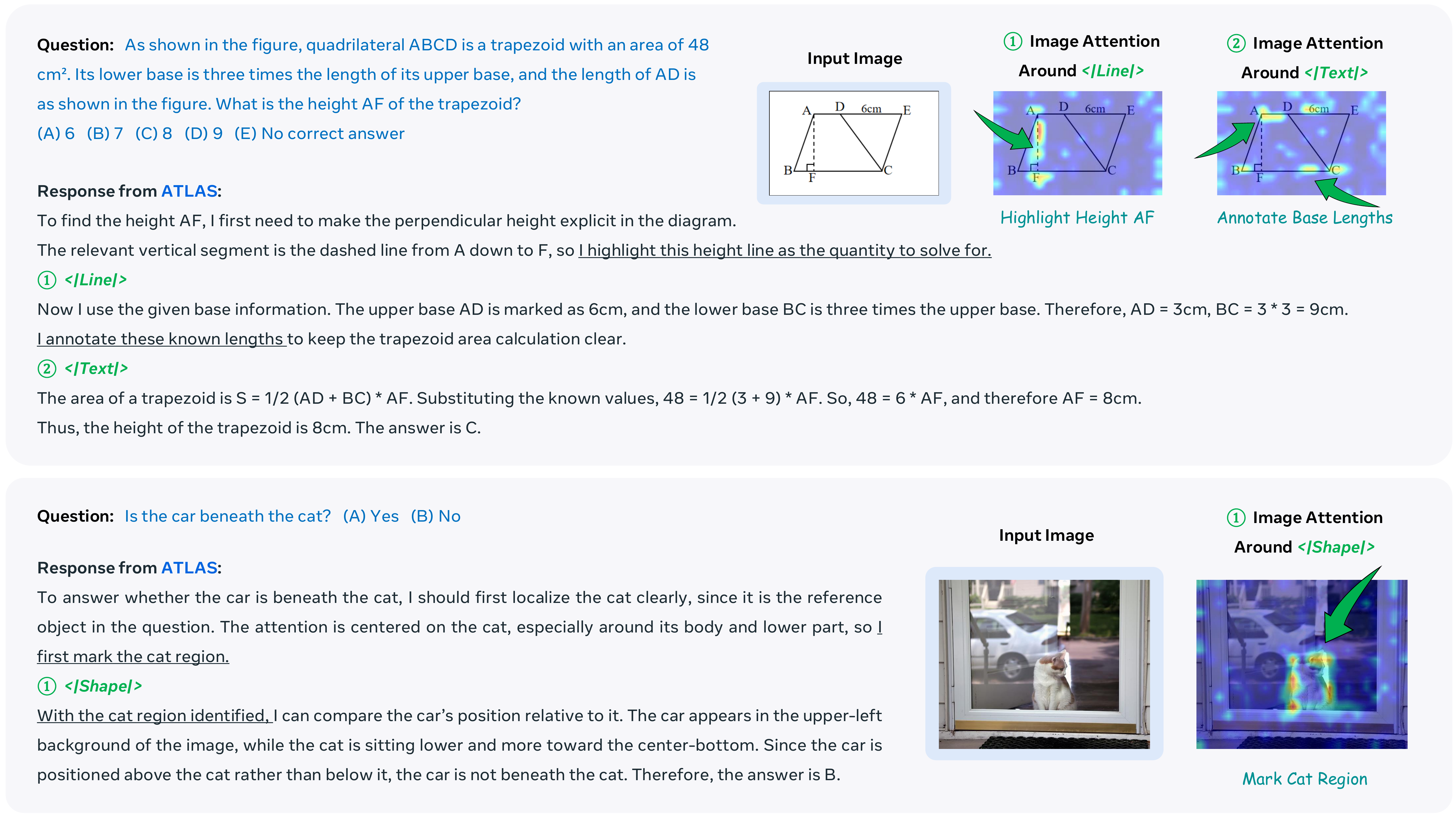}
    \caption{\textbf{Attention Analysis of Functional Tokens.} The highlighted regions show that functional tokens tend to attend to task-relevant visual evidence like geometric lines and target objects.}
    \label{fig:attention_analysis} 
\end{figure*}

\begin{figure*}[t!]
    \centering
    \includegraphics[width=\textwidth]{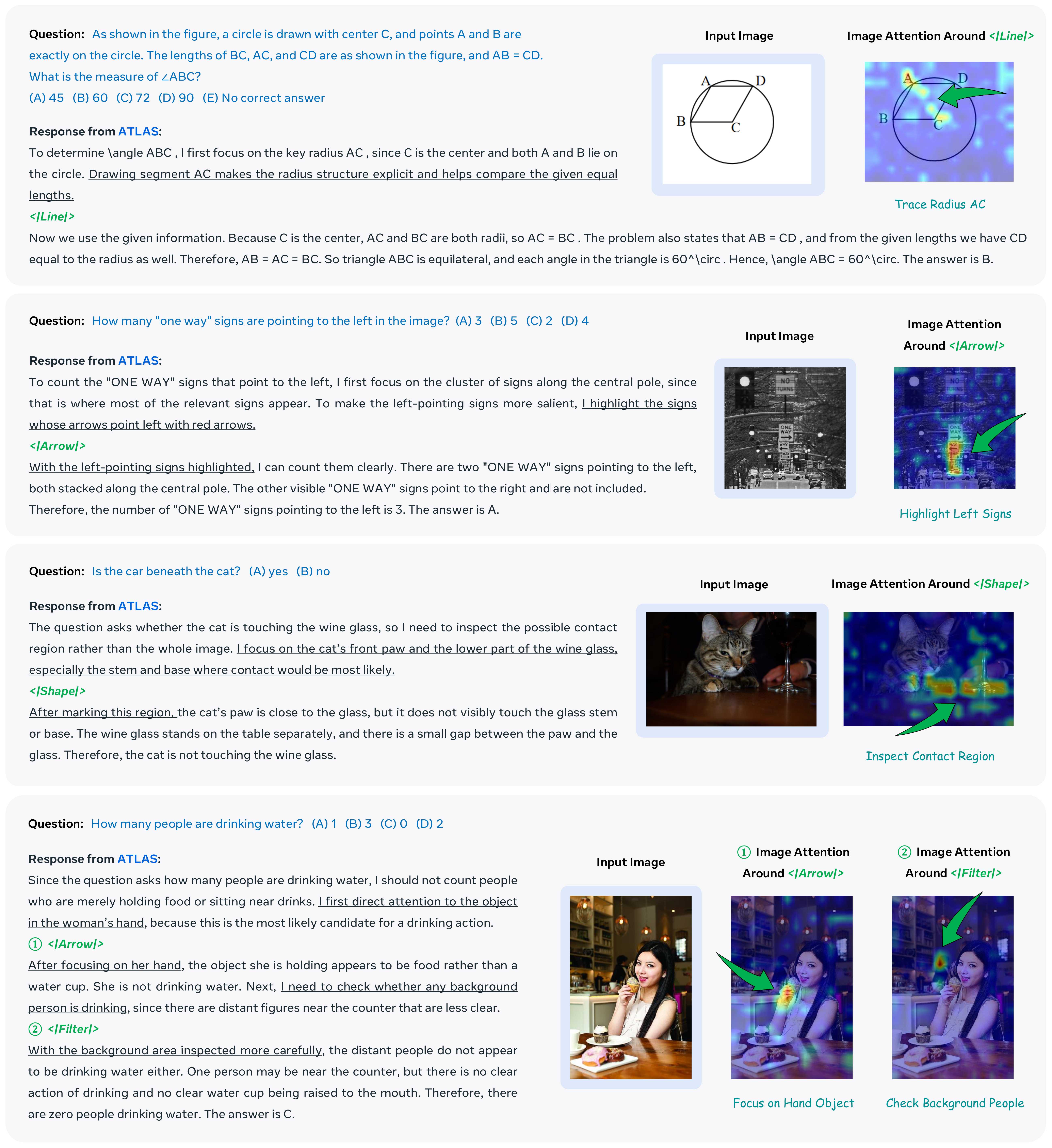}
    \caption{\textbf{Attention Visualizations around Functional Tokens.} The attention maps show that functional tokens tend to focus on task-relevant visual evidence across different examples.}
    \label{fig:additional_attention}
\end{figure*}

\subsection{Quantitative Analysis}
\label{sec:quantitative_analysis}
The performance comparison across visual reasoning benchmarks is summarized in Tab.~\ref{tab:benchmark_results}. Compared with the base model Qwen2.5-VL~\cite{bai2025qwen2}, \name\ brings clear improvements across all benchmarks. The improvement is notable on BLINK, where Qwen2.5-VL achieves an average accuracy of 22.8\%, while ATLAS$_{\text{\textit{LA-GRPO}}}$ reaches 51.3\%. This shows that discrete vocabulary functional tokens can effectively enhance structured visual reasoning.

Among our variants, ATLAS$_{\text{\textit{SFT}}}$ already improves the BLINK average from 22.8\% to 46.0\%, demonstrating that supervised fine-tuning provides a strong reasoning capability. ATLAS$_{\text{\textit{GRPO}}}$ further improves the overall performance, reaching 50.5\% on BLINK and 40.3\% on WeMath. It brings large gains on several subsets, such as Jigsaw and Spatial Relation, but the improvements are not uniform. For example, it drops on IQ and multi-view reasoning compared with ATLAS$_{\text{\textit{SFT}}}$. This suggests that sequence-level preference optimization can improve final-answer accuracy, but may also introduce unstable functional-token behavior on structured reasoning tasks. With LA-GRPO, \name\ achieves the best performance among our variants on WeMath and BLINK average, reaching 45.0\% and 51.3\%, respectively. It also improves over standard GRPO on several subsets, including Art Style, Counting, Forensic Detection, and Multi-view Reasoning, with multi-view reasoning increasing from 43.6\% to 53.4\%. Compared with existing visual reasoning methods, ATLAS$_{\text{\textit{LA-GRPO}}}$ obtains consistently competitive results. Although standard GRPO performs better on a few metrics such as V*, Jigsaw, and Spatial Relation, LA-GRPO gives a more balanced result across the benchmark. These results indicate that anchoring the optimization on functional tokens helps stabilize training and improves the overall effectiveness of \name\ on complex visual reasoning tasks.

\subsection{Qualitative Analysis}
\label{sec:qualitative_analysis}

\paragraph{Qualitative Examples.}
Fig.~\ref{fig:qualitative_analysis} and Fig.~\ref{fig:additional_qualitative} show representative visual reasoning trajectories from \name. As shown, the model invokes functional tokens at meaningful steps: \texttt{<|Shape|>} is used to localize relevant regions, \texttt{<|Arrow|>} guides attention to additional evidence, and \texttt{<|Text|>} supports counting and labeling. These examples show that functional tokens are not used as isolated markers, but are naturally integrated into the reasoning process to support visual reasoning.

\paragraph{Attention Scores.}
We further visualize the image attention around functional tokens in Fig.~\ref{fig:attention_analysis} and Fig.~\ref{fig:additional_attention}. For each functional token, we average the attention scores between image tokens and the nearby 10 tokens around it. The resulting maps show that different functional tokens tend to focus on relevant visual regions, \textit{e.g.}, in Fig.~\ref{fig:attention_analysis}, \texttt{<|Line|>} attends to the height segment in the geometry problem, and \texttt{<|Shape|>} highlights the cat region for spatial comparison. These patterns suggest that functional tokens are associated with meaningful visual evidence rather than being used only as textual markers.

\subsection{Efficiency Analysis}
\label{sec:efficiency_analysis}

We further evaluate the inference efficiency of \name on BLINK-Jigsaw~\cite{fu2024blink} and compare it with V-Thinker~\cite{qiao2025v}, which relies on explicit agentic reasoning. As shown in Tab.~\ref{tab:efficiency}, \name greatly reduces both output length and operation-formulation overhead. This is because each visual operation is represented by a single functional token, rather than a long textual description or tool-call formulation. This compact formulation directly improves inference efficiency. \name reduces the average latency from 18.83\textit{s} to 3.80\textit{s} and lowers peak memory usage from 2.55GB to 1.43GB, while maintaining competitive accuracy. These results show that representing visual operations as compact functional tokens can reduce inference cost while preserving effective visual reasoning.

\begin{table}[t!]
\centering
\caption{\textbf{Efficiency Comparison.} Efficiency metrics are reported as per-query averages. \name\ substantially reduces generation overhead and latency while improving accuracy.}
\label{tab:efficiency}
\renewcommand{\arraystretch}{1.12}
\resizebox{0.8\linewidth}{!}{
\begin{tabular}{l|ccccc|c}
\toprule
\textbf{Method} 
& \textbf{All Tokens $\downarrow$} 
& \textbf{Code Tokens $\downarrow$} 
& \textbf{Func. Tokens $\downarrow$} 
& \textbf{Latency $\downarrow$} 
& \textbf{Peak Mem. $\downarrow$} 
& \textbf{Acc. $\uparrow$} \\
\cmidrule(lr){1-1} \cmidrule(lr){2-7}
V-Thinker~\citep{qiao2025v} 
& 489.57 
& 350.35 & -
& 18.83s 
& 2.55GB 
& 42.0 \\
{\textcolor{metablue}{ATLAS}}
& {99.85} 
& -
& {0.81} 
& {3.80s} 
& {1.43GB} 
& {57.7} \\
\cmidrule(lr){1-1} \cmidrule(lr){2-7}
\textcolor{blue}{\textit{Relative Gain} }
& \textcolor{blue}{\textit{4.90$\times$} }
& \multicolumn{2}{c}{\textcolor{blue}{\textit{434.3$\times$}}}
& \textcolor{blue}{\textit{4.96$\times$} }
& \textcolor{blue}{\textit{1.78$\times$} }
& \textcolor{blue}{\textit{+15.7}} \\
\bottomrule
\end{tabular}
}
\end{table}

\subsection{Ablation Study of Negative Reward Penalties}
\label{sec:appendix_reward_ablation}
We ablate the format reward ($r_{\text{fmt}}$), length penalty ($p_{\text{len}}$), and token spam penalty ($p_{\text{spam}}$) to verify the necessity of each reward constraint. As shown in Table~\ref{tab:reward_penalty_ablation}, removing any of these components leads to a performance drop on BLINK, indicating that the negative reward design is important for stable functional-token alignment. The full LA-GRPO objective achieves the best BLINK average accuracy of 51.3. When the format reward is removed, the model becomes less reliable in producing parseable final answers, leading to a drop of 1.3. Removing the length penalty causes responses to become unnecessarily verbose, which increases generation cost and hurts reasoning accuracy. The largest degradation occurs when removing the token spam penalty, where the BLINK average drops from 51.3 to 47.0. In this setting, we observe severe reward hacking: the model generates up to 18.7 functional tokens per sequence merely to accumulate $r_{\text{func}}$, rather than using them to support effective reasoning. Similarly, removing $p_{\text{len}}$ increases the average sequence length by 43.8\%, resulting in higher compute cost without accuracy gains. These results show that the format, length, and token-spam constraints are complementary: they encourage \name to invoke functional tokens only when useful, while keeping the reasoning trajectory concise, well formatted, and robust.

\begin{table}[t!]
\centering 
\caption{\textbf{Ablation Study of Negative Reward Penalties in LA-GRPO on BLINK.} Removing any component leads to lower BLINK average accuracy.}
\label{tab:reward_penalty_ablation}
\begin{tabular}{cccc}
\toprule
$r_{\text{fmt}}$ & $p_{\text{len}}$ & $p_{\text{spam}}$ & Acc. \\
\midrule
\checkmark & \checkmark & \checkmark & 51.3 \\
-          & \checkmark & \checkmark & 50.0 \\
\checkmark & -          & \checkmark & 49.2 \\
\checkmark & \checkmark & -          & 47.0 \\
\bottomrule
\end{tabular}
\end{table}
\section{Related Work}
\label{sec:related}

\paragraph{Agentic Visual Reasoning.}
Another line of research equips language or multi-modal models with the ability to act through external tools. Representative examples include program-based and code-based systems such as VISPROG~\cite{gupta2023visual} and ViperGPT~\cite{suris2023vipergpt}, where the model generates executable programs to invoke specialized vision modules. More recent agentic frameworks broaden this idea by integrating richer tool ecosystems, hierarchical planning, and mixed-modality execution~\cite{zheng2025deepeyes,qiao2025v,shao2024visual,wang2025pixel}. Related work also explores visual workspaces such as sketchpads, where the model produces intermediate drawings, marks, or auxiliary constructions to support subsequent reasoning~\cite{hu2024visual}. These methods are effective because they allow the model to actively manipulate visual inputs rather than passively observe them once. However, the underlying execution typically happens through external programs, APIs, or auxiliary environments. This creates a disjoint reasoning loop in which visual action is performed outside the standard autoregressive computation graph. Consequently, such approaches are usually non-differentiable end-to-end, incur latency from context switching and tool execution, and often require verbose code or program generation even for relatively simple visual operations.

\paragraph{Latent Visual Reasoning.}
Latent reasoning offers a promising alternative by moving intermediate computation from explicit text into compact hidden representations~\cite{zhang2023multimodal,hao2024training}. In language modeling, recent work explores this direction from several perspectives, including self-generated latent rationales~\cite{zelikman2024quiet}, continuous thought representations~\cite{hao2024training}, and recurrent-depth architectures that scale test-time computation without emitting long reasoning traces~\cite{zhao2025learning}. In multi-modal settings, methods such as Heima~\cite{shen2025efficient} compress explicit reasoning into hidden thinking tokens, while more recent approaches introduce latent visual tokens or latent visual trajectories to support multi-modal reasoning without full image generation~\cite{li2025latent, qin2025chain, wang2025monet}. These studies show that latent reasoning can improve efficiency and, in some cases, performance. Nevertheless, existing methods still face important limitations. Many rely on auxiliary supervision, reconstruction, or distillation targets for latent states~\cite{qin2025chain,wang2025monet}, which restricts flexibility and may limit generalization beyond the training setup. More importantly, several approaches introduce recurrent or non-standard computation patterns, which deviate from the standard next-token prediction pipeline and reduce compatibility with highly optimized autoregressive parallel training systems. In contrast, our method formulates agentic visual actions as discrete functional tokens within the normal vocabulary space, keeping the entire reasoning process strictly inside the standard autoregressive loop. This preserves end-to-end differentiability while avoiding the recurrent dependencies and external execution overhead that limit prior approaches.

\section{Conclusion}
In this paper, we introduced \textcolor{metablue}{\textbf{\name}}, a visual reasoning framework that represents visual operations as discrete functional tokens within the standard autoregressive vocabulary. By internalizing visual reasoning into compact tokens, \name avoids intermediate image generation, external tool execution, and verbose operation formulations, while preserving interpretability and compatibility with parallel autoregressive training. We further identified gradient dilution for sparse functional tokens during GRPO training and proposed Latent-Anchored GRPO to stabilize their optimization. Extensive experiments show that \name establishes a distinct visual reasoning paradigm, achieving strong performance on complex benchmarks with reduced inference latency and memory usage.

\newpage
\bibliographystyle{assets/plainnat}
\bibliography{paper}
\end{document}